\DocumentMetadata 
{
	lang		= en-US,
	pdfversion  = 1.7,
    pdfstandard = a-2b,
}

\documentclass[letterpaper, 10 pt]{ieeeconf}

\usepackage{amsfonts}
\usepackage{amsmath}
\usepackage{amssymb}

\usepackage{listings}
\usepackage{prettyref}
\usepackage{xspace}
\usepackage{xcolor}
\usepackage{graphicx}
\usepackage{bm}

\usepackage{multirow}
\usepackage{hhline}
\usepackage{booktabs}
\usepackage{adjustbox}

\usepackage{subfig}
\usepackage{floatrow}
\usepackage{tikz}
\usepackage{pgfplots}
\usetikzlibrary{fillbetween}
\usepgfplotslibrary{fillbetween}
\pgfplotsset{compat=1.18}

\usepackage[short]{optidef}

\usepackage{algorithm2e}
\usepackage[font=small,labelfont=bf]{caption}
\SetAlCapNameFnt{\small}
\SetAlCapFnt{\small}

 \usepackage[letterpaper,
            left=1in,
            right=1in,
            top=1in,
            bottom=1in]{geometry}
\usepackage{hyperref}

\graphicspath{ {figures/}}

\newrefformat{prob}{Problem\,\ref{#1}}
\newrefformat{def}{Definition\,\ref{#1}}
\newrefformat{sec}{Section\,\ref{#1}}
\newrefformat{sub}{Section\,\ref{#1}}
\newrefformat{prop}{Proposition\,\ref{#1}}
\newrefformat{app}{Appendix\,\ref{#1}}
\newrefformat{alg}{Algorithm\,\ref{#1}}
\newrefformat{cor}{Corollary\,\ref{#1}}
\newrefformat{thm}{Theorem\,\ref{#1}}
\newrefformat{lem}{Lemma\,\ref{#1}}
\newrefformat{fig}{Fig.\,\ref{#1}}
\newrefformat{tab}{Table\,\ref{#1}}

\newtheorem{theorem}{Theorem}
\newtheorem{problem}{Problem}

\newtheorem{lemma}[theorem]{Lemma}

\newtheorem{proposition}[theorem]{Proposition}

\newcommand{\bdmath}{\begin{dmath}}
\newcommand{\edmath}{\end{dmath}}
\newcommand{\beq}{\begin{equation}}
\newcommand{\eeq}{\end{equation}}
\newcommand{\bdm}{\begin{displaymath}}
\newcommand{\edm}{\end{displaymath}}
\newcommand{\bea}{\begin{eqnarray}}
\newcommand{\eea}{\end{eqnarray}}
\newcommand{\beal}{\beq \begin{array}{ll}}
\newcommand{\eeal}{\end{array} \eeq}
\newcommand{\beas}{\begin{eqnarray*}}
\newcommand{\eeas}{\end{eqnarray*}}
\newcommand{\ba}{\begin{array}}
\newcommand{\ea}{\end{array}}
\newcommand{\bit}{\begin{itemize}}
\newcommand{\eit}{\end{itemize}}
\newcommand{\ben}{\begin{enumerate}}
\newcommand{\een}{\end{enumerate}}

\newcommand{\eg}{\emph{e.g.,}\xspace}
\newcommand{\ie}{\emph{i.e.,}\xspace}

\newcommand{\hide}[1]{}

\newcommand{\hiddenText}{{\color{gray} hidden text.}}
\newcommand{\hideWithText}[1]{\hiddenText}

\DeclareMathOperator*{\argmin}{arg\,min}

\newcommand{\eye}{{\mathbf I}}

\newcommand{\at}[1]{^{(#1)}}

\newcommand{\SO}[1]{\ensuremath{\mathrm{SO}(#1)}\xspace}

\newcommand{\T}{\mathsf{T}}

\newcommand{\blue}[1]{{\color{blue}#1}}

\newcommand{\linkToPdf}[1]{\href{#1}{\blue{(pdf)}}}
\newcommand{\linkToPpt}[1]{\href{#1}{\blue{(ppt)}}}
\newcommand{\linkToCode}[1]{\href{#1}{\blue{(code)}}}
\newcommand{\linkToWeb}[1]{\href{#1}{\blue{(web)}}}
\newcommand{\linkToVideo}[1]{\href{#1}{\blue{(video)}}}
\newcommand{\linkToMedia}[1]{\href{#1}{\blue{(media)}}}
\newcommand{\award}[1]{\xspace}

\newcommand{\eps}{\epsilon}
\newcommand{\RR}{\mathbb{R}}

\newrefformat{appendix}{Appendix\ \ref{#1}}

\def\*#1{\mathbf{#1}}
\def\'#1{\bm{#1}}

\newcommand{\minproblem}[3]{
    \min_{#1} \: & #2 \\
    \textrm{s.t.} \quad & #3
}

\newcommand{\scfstar}{$\text{SCF}^\star$}
\newcommand{\pace}{$\text{PACE}^\star$}

\title{\huge Category-Level Object Shape and Pose Estimation \\ in Less Than a Millisecond}

\author{Lorenzo Shaikewitz$^1$, Tim Nguyen$^2$, and Luca Carlone$^1$
    \thanks{This work was supported by the AFOSR “Certifiable and Self-Supervised Category-Level Tracking” program, Carlone’s NSF CAREER award, and the ONR RAPID program. L. Shaikewitz is supported by an NSF graduate research fellowship. L. Carlone holds concurrent appointments at MIT and as an Amazon Scholar. This paper describes work performed at MIT and is not associated with Amazon.}
    \thanks{$^{1}$L. Shaikewitz and L. Carlone are with the Laboratory for Information and Decision Systems, Massachusetts Institute of Technology, Cambridge, MA.
            Emails: {\tt\footnotesize \{lorenzos, lcarlone\}@mit.edu}.}
    \thanks{$^{2}$T. Nguyen is with Boston University, Boston, MA. Work was completed during an internship at MIT.
            {\tt\footnotesize timnguyen737@gmail.com}.}
}

\begin{document}

\maketitle
\thispagestyle{empty}
\pagestyle{empty}

\begin{tikzpicture}[overlay, remember picture]
\node[anchor=north, yshift=-0.5cm, text width=2\textwidth, align=center] at (current page.north) {
    \textbf{This paper has been accepted for publication in the \emph{2026 IEEE International Conference on Robotics and Automation.}}\\
    Please cite the paper as: Lorenzo Shaikewitz, Tim Nguyen, and Luca Carlone,\\
    ``Category-Level Object Shape and Pose Estimation in Less Than a Millisecond,''\\
    in \emph{2026 IEEE International Conference on Robotics and Automation (ICRA)}
};
\end{tikzpicture}

\begin{abstract}
    Object shape and pose estimation is a foundational robotics problem, supporting tasks from manipulation to scene understanding and navigation. We present a fast local solver for shape and pose estimation which requires only category-level object priors and admits an efficient \emph{certificate} of global optimality. Given an RGB-D image of an object, we use a learned front-end to detect sparse, category-level semantic \emph{keypoints} on the target object. We represent the target object's unknown shape using a linear \emph{active shape model} and pose a maximum a posteriori optimization problem to solve for position, orientation, and shape simultaneously. Expressed in unit quaternions, this problem admits first-order optimality conditions in the form of an eigenvalue problem with eigenvector nonlinearities. Our primary contribution is to solve this problem efficiently with \emph{self-consistent field iteration}, which only requires computing a $4\times4$ matrix and finding its minimum eigenvalue-vector pair at each iterate. Solving a linear system for the corresponding Lagrange multipliers gives a simple global optimality certificate. One iteration of our solver runs in about 100 microseconds, enabling fast outlier rejection. We test our method on synthetic data and a variety of real-world settings, including two public datasets and a drone tracking scenario. Code is released at \url{https://github.com/MIT-SPARK/Fast-ShapeAndPose}.
\end{abstract} 

\section{Introduction}
\label{sec:intro}

A diverse set of robotics applications benefits from object shape and pose estimation. Autonomous cars, for example, need to locate obstacles and other cars~\cite{Koopman16sae-autonomousVehicleTesting}, while household manipulators need to locate objects to interact with~\cite{Tremblay18corl-DeepObject}. 
In many of these applications the object shape is not known exactly but its \emph{category} is available (\eg from a semantic segmentation method). We consider this setting and derive a shape and pose estimator using category-level priors.

The work of Shi et al.~\cite{Shi23tro-PACE} established a certifiably optimal approach for category-level shape and pose estimation using a semidefinite relaxation. We consider a similar problem setup but emphasize both \emph{speed} and \emph{certifiability}. A fast estimator allows quick reaction to new inputs, performance with limited compute, and comprehensive outlier rejection~\cite{Yang20ral-GNC,Fischler81}. Certifiability provides an \emph{a posteriori} guarantee that the estimate returned is statistically optimal. When the certificate fails, the user can decide to trust the output, try a different initialization, or acquire a new batch of measurements.

\begin{figure}[tb]
    \centerline{\includegraphics[width=\linewidth]{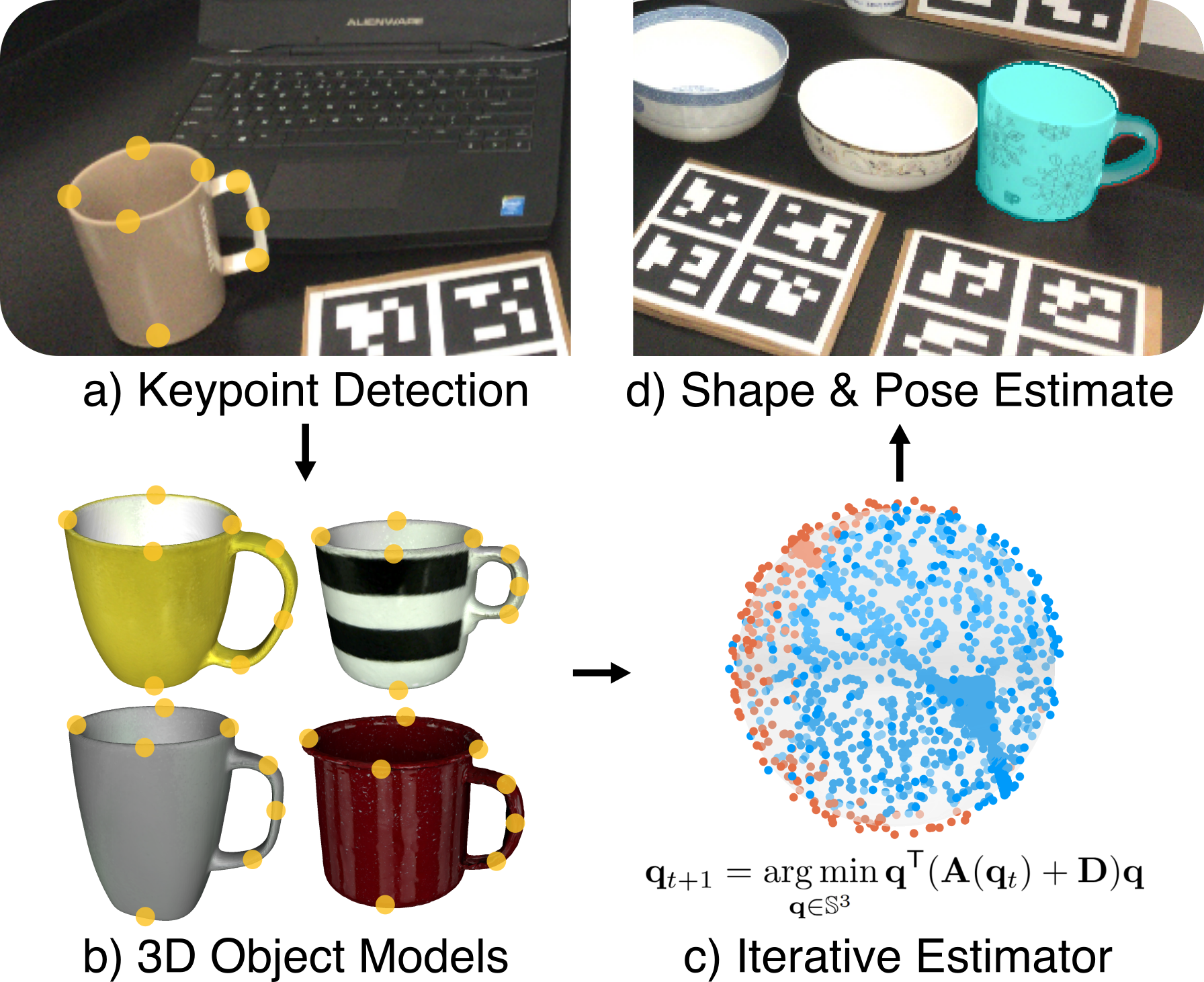}}
    \caption{\textbf{Overview of Method.} Given 3D keypoint detections (a), on an RGB-D image, and a category-level shape library (b), we use self-consistent field iteration (c), to estimate the shape and pose of an object (d).
    \vspace{-15pt}
    }
    \label{fig:intro}
\end{figure}

Our algorithm relies on the eigenvalue structure of the first-order optimality conditions written in the quaternion representation of rotations. It returns local solutions which are often globally optimal. To verify this, we introduce a fast global optimality certifier based on Lagrangian duality. Specifically, our contributions are:
\begin{itemize}
    \item A fast local solver for category-level shape and pose estimation 
    using self-consistent field iteration~\cite{Cai18-nepv}.
\item A fast \emph{a posteriori} certificate of global optimality for our local solutions.
    \item Experimental evaluation of runtime and accuracy on synthetic data, a drone tracking scenario, and two large-scale datasets.
\end{itemize}

The remainder of the paper is organized as follows. We begin with a literature review (\prettyref{sec:related}) and quaternion preliminaries (\prettyref{sec:quat}). Then, we give the problem formulation in \prettyref{sec:problem} and reformulate it with quaternions in \prettyref{sec:nepv}. 
To solve the resulting nonlinear eigenproblem, we use self-consistent field iteration for local solutions and SDP optimality conditions to certify global optimality in \prettyref{sec:scf}. In \prettyref{sec:experiments}, we show our method is significantly faster than other local solvers and learned baselines. \section{Related Work}
\label{sec:related}

\textbf{Solvers for Rotation Estimation Problems.}
Rotation estimation encompasses a large class of non-convex problems beyond pose estimation. When posed in a least-squares form, they can be solved with Gauss-Newton~\cite{Gauss1809} or Levenberg-Marquardt~\cite{Levenberg44qam}, which approximate Newton's method by linearizing the residuals. A more specialized set of approaches use the manifold structure of $\SO3$ and Riemannian counterparts to Gauss-Newton or gradient descent~\cite{manopt, Hertzberg08-manifolds, Chen22cvpr-manifoldPGD,Smith94ams}. These approaches generate \emph{local} solutions. Global optimality is largely achieved via a posteriori optimality \emph{certificates} using Shor's semidefinite relaxation~\cite{Shor87} or the Burer-Monteiro method~\cite{Burer03mp}. This approach has been applied to a variety of rotation estimation problems including rotation averaging~\cite{Brynte21-semidefiniteRotation}, pose graph optimization~\cite{Rosen18ijrr-sesync}, and pose estimation~\cite{Shi23tro-PACE,Shaikewitz24ral-CAST}.

A notable exception to these paradigms is point cloud registration, which can be solved globally via eigenvalue decomposition \cite{Arun87pami,Shuster89jas}. In this paper, we show the shape and pose estimation problem admits a similar (albeit nonlinear) eigenvalue structure, and leverage that structure to construct a fast certifiable solver.

\textbf{Category-Level Shape and Pose Estimation.}
Category-level shape and pose estimation largely reduces to correspondence and alignment stages. At the most explicit,~\cite{Shi23tro-PACE} uses a learned keypoint detector~\cite{Ke20-gsnet} to estimate pixel-wise correspondences and optimize for the maximum likelihood object shape and pose. Other methods use local features~\cite{Wen21iros-bundletrack, Liu22eccv-CATRE} or semantic features~\cite{Chen24cvpr-secondpose} to build and align a shape estimate. To avoid explicit features, normalized object coordinates~\cite{Wang19-normalizedCoordinate} compute pixel-wise correspondences in a normalized frame and then regress for object pose~\cite{Shi25cvpr-CRISP,Tian20eccv-SPD,Wang19-normalizedCoordinate}. We refer the reader to~\cite{Liu24arxiv-deepLearningSurvey} for a comprehensive survey. In this paper, we adopt the problem setup from~\cite{Shi23tro-PACE}, using a keypoint detector and focusing on the alignment stage.
 \section{Preliminaries on Quaternion Arithmetic}
\label{sec:quat}

In this section, we review quaternion arithmetic for rigid rotations (for more detail, see~\cite{Altmann13-Rotations}). A rotation about axis $\'\omega\in\RR^3$ by angle $\theta\in\RR$ admits the following representation as a unit quaternion $\*q$:
\begin{equation}
    \label{eq:quataxang}
    \*q = \begin{bmatrix}
        \cos(\theta/2)\\
        \'\omega\sin(\theta/2)
    \end{bmatrix}\!.
\end{equation}

From this definition, it is clear that negating the last three elements (the \emph{vector part} of the quaternion) gives the inverse rotation: $\*q^{-1} = [q_1, -\*q_v^\T]^\T$. The first element $q_1$ is called the \emph{scalar part}. We also observe that there are two unit quaternions for every rigid rotation: $-\*q$ and $\*q$ represent the same rotation. To rotate a point $\*y\in\RR^3$, we use a \emph{quaternion product} $\circ$. Specifically,
\begin{equation}
    \label{eq:quatprodrot}
    \*q\circ
    \begin{bmatrix}
        0\\\*y
    \end{bmatrix}
    \circ\*q^{-1}
     =
     \begin{bmatrix}
        0\\
        \*R\*y
     \end{bmatrix}\!,
\end{equation}
where $\*R\in\SO3$ is the rotation matrix corresponding to the quaternion $\*q$. Quaternion products may be written as matrix-vector products. For quaternions $\*a, \*b\in\RR^4$,
\begin{equation}
    \label{eq:quatmath}
    \*a \circ \*b = \*\Omega_l(\*a)\*b = \*\Omega_r(\*b)\*a.
\end{equation}
In~\eqref{eq:quatmath}, $\*\Omega_l$ and $\*\Omega_r$ are the product matrices:
\begin{equation}
    \label{eq:omegas}
    \resizebox{0.9\columnwidth}{!}
    {$\displaystyle
    \*\Omega_l(\*a) \triangleq
    \begin{bmatrix}
        a_1 & -a_2 & -a_3 & -a_4\\
        a_2 & a_1 & -a_4 & a_3 \\
        a_3 & a_4 & a_1 & -a_2\\
        a_4 & -a_3 & a_2 & a_1
    \end{bmatrix}
    \!\text{, }
    \*\Omega_r(\*a) \triangleq
    \begin{bmatrix}
        a_1 & -a_2 & -a_3 & -a_4\\
        a_2 & a_1 & a_4 & -a_3 \\
        a_3 & -a_4 & a_1 & a_2\\
        a_4 & a_3 & -a_2 & a_1
    \end{bmatrix}\!.
    $}
\end{equation}

In this paper we allow $\*\Omega_l$ and $\*\Omega_r$ to take vectors $\*y\in\RR^3$ by implicitly homogenizing them with a leading $0$. A little algebra shows that a Euclidean inner product involving a rotation matrix $\*R$ can be written as a quadratic form in a corresponding quaternion $\*q$.
\begin{lemma}[\cite{Yang19iccv-QUASAR}]
    \label{lem:quatidentity}
    Let the unit quaternion $\*q\in\mathbb{S}^3$ represent the same rotation as the matrix $\*R\in\SO{3}$. For $\*x, \*y\in\RR^3$ vectors: $\*x^\T \*R \*y = -\*q^\T \*\Omega_l(\*x)\*\Omega_r(\*y)\*q$.
\end{lemma} 
\section{Category-Level Shape and Pose Estimation Problem}
\label{sec:problem}
Given detections of 3D keypoints on an object of known category, we estimate its shape and pose (position and orientation). This section describes the problem formulation, including our choice of shape representation and measurement model. We adopt the same problem as~\cite{Shi23tro-PACE}, rephrased here for clarity.

\subsection{Active Shape Model}
For each category, we assume a library of $K$ representative 3D shapes (point clouds) that span the category according to a linear \emph{active shape model}. For each point $\*x_i\in\RR^3$ on an arbitrary object in the category, $\*x_i$ may be expressed as a linear combination of corresponding points $\*b_k^i\in\RR^3$ on the objects in the 3D shape library. Mathematically:
\begin{equation}
    \label{eq:scfasm}
    \*x_i = \sum_{k=1}^K c_k \*b_{k}^i \triangleq \*B_i\*c,
\end{equation}
where $\*B_i\in\RR^{3\times K}$ stacks each $\*b_{k}^i$ as columns and $\*c$ defines a linear combination: $c_k\in[0,1]$ and $\sum_{k=1}^K c_k = 1$. It is useful to think of these points as semantically related. For example, within the \emph{bottle} category, a point on each shape could be the center of its bottle cap. The active shape model can represent any object in the convex hull of its 3D shape library. For more than two objects this becomes quite expressive; in the bottle example, the active shape model could represent any cap between the shortest and tallest in the library.

\subsection{Measurement Model}
Given an object's category, we estimate its shape and pose from a sparse set of 3D \emph{keypoints} $\*y_i\in\RR^3$, $i=1,...,N$, with known associations to a point in each of the library shapes $\*B_i$. These measurements may come from pixel detections by a learned keypoint detector~\cite{He17iccv-maskRCNN} combined with depth information, and are typically semantically meaningful (see~\prettyref{fig:intro}).

Let $\*c$ be the object's shape vector. Denoting the object's position $\*p\in\RR^3$ and orientation $\*R\in\SO{3}$ with respect to some fixed reference frame (\ie the camera frame), the detected keypoints $\*y_i$, $i=1,...,N$, obey the following generative model:
\begin{equation}
    \label{eq:pacemeas}
    \*y_i = \*R\*B_i\*c + \*p + \'\eps_i.
\end{equation}

Eq.~\eqref{eq:pacemeas} models the keypoint measurement as a linear combination of shape library points that is rotated and translated before being perturbed by measurement noise $\'\eps_i$. We assume the measurement noise follows an isotropic Gaussian distribution with zero mean and known covariance: $\'\eps_i\sim\mathcal{N}(0,w_i^{-1}\eye_3)$. Our goal is to estimate the object pose and shape vector from noisy measurements. This is the \emph{category-level shape and pose estimation} problem.
\begin{problem}
    \label{prob:pace}
    Estimate the shape $\*c$ and pose $(\*R,\*p)$ of an object from $N$ 3D keypoint measurements with known category-level associations.
\end{problem} \section{Nonlinear Eigenproblem for Local Solutions}
\label{sec:nepv}
In this section, we rephrase~\prettyref{prob:pace} as a non-convex optimization problem and develop its first-order optimality conditions into a nonlinear eigenvalue problem. 
We recall from~\cite{Shi23tro-PACE} that the optimal position and shape solving~\prettyref{prob:pace} can be computed in closed form, leading to a rotation-only maximum a posteriori (MAP) estimation problem. While~\cite{Shi23tro-PACE} solves this non-convex problem with a semidefinite relaxation, we re-express the rotation estimation problem with quaternions and show its first-order optimality conditions form a nonlinear eigenproblem. Our solver exploits this structure.

Under model~\eqref{eq:pacemeas} with Gaussian noise and a shape prior $\*c \sim \mathcal{N}(\*0, \lambda^{-1}\*I_K)$, the MAP estimator solving \prettyref{prob:pace} takes the following form~\cite{Shi23tro-PACE}:
\begin{equation}
\begin{aligned}
    \label{eq:pace_original}
    \minproblem{\substack{
        \*R\in\SO3\\
        \*p\in\RR^3,\:\*c\in\RR^K
    }}
    {\sum_{i=1}^N w_i\|\*y_i - \*R\*B_i\*c - \*p\|^2 + \lambda\|\*c\|^2}
    {\*1^\T\*c = 1,\:\*c\in[0,1]^K.}
\end{aligned}
\end{equation}

Following~\cite{Shi23tro-PACE}, we drop the constraint $\*c\in[0,1]^K$ for the remainder of the paper. While~\eqref{eq:pace_original} has a convex quadratic objective, the constraint $\*R\in\SO3$ introduces non-convexity~\cite{Rosen18ijrr-sesync}. Fortunately,~\eqref{eq:pace_original} is convex in $\*p$ and $\*c$ as a function of $\*R$. This allows analytic elimination of the shape and position variables via the first-order optimality conditions, summarized below.

\begin{proposition}[Optimal Shape and Position~\cite{Shi23tro-PACE}]
    Given a rotation estimate $\*R$ and shape vector $\*c$, the optimal position solving \eqref{eq:pace_original} is:
    \begin{equation}
        \label{eq:paceoptp}
        \*p^\star(\*R, \*c) = \*{\bar{y}} - \*R\*{\bar{B}}\*c,
    \end{equation}
    where $\*{\bar{y}}$ and $\*{\bar{B}}$ are weighted averages of $\*y_i$ and $\*B_i$:
    \begin{equation}
        \*{\bar{y}} \triangleq \frac{\sum_{i=1}^N w_i \*y_i}{\sum_{i=1}^N w_i}
        \quad\text{and}\quad \*{\bar{B}} \triangleq \frac{\sum_{i=1}^N w_i \*B_i}{\sum_{i=1}^N w_i}.
    \end{equation}
    The optimal shape vector solving~\eqref{eq:pace_original} can be recovered from a rotation estimate:
    \begin{equation}
        \label{eq:paceoptc}
        \*c^\star(\*R) = \*C_1 \sum_{i=1}^N \left(\*{\bar{B}}_i^\T \*R^\T \*{\bar{y}}_i\right) + \*c_2,
    \end{equation}
    where we use the following symbols:
    \begin{equation}
        \begin{array}{ll}
            \*{\hat{B}}^2 \hspace{-7pt}&\triangleq \sum_{i=1}^N \*{\bar{B}}_i^\T \*{\bar{B}}_i,\\
            \*H \hspace{-7pt}&\triangleq \*{\hat{B}}^2 + \lambda\*I_K,\\
            \*C_1 \hspace{-7pt}&\triangleq 
                \*H^{-1} - \*H^{-1}\*1_K\left(
                    \*1_K^\T\*H^{-1}\*1_K
                \right)^{-1}\*1_K^\T \*H^{-1},\\
            \*c_2 \hspace{-7pt}&\triangleq \*H^{-1}\*1_K\left(
                \*1_K^\T\*H^{-1}\*1_K
            \right)^{-1}.
        \end{array}
    \end{equation}
    Note that $\*{\hat{B}}^2$ is invertible as long as there are $N\geq 3$ non-colinear keypoints and $N \geq K$. The $\lambda$ term regularizes the problem to ensure invertibility when the latter condition is violated, \ie when $N < K$.
\end{proposition}

Substituting the optimal position \eqref{eq:paceoptp} and shape \eqref{eq:paceoptc} into \eqref{eq:pace_original}, we rephrase object shape and pose estimation as a rotation-only estimation problem.

\begin{problem}
    \label{prob:rotest}
    Estimate the object's MAP rotation:
    \begin{equation}
        \label{eq:paceqcqp_rotonly}
        \min_{\*R\in\SO3} \sum_{i=1}^N\left\|\*{\bar{y}}_i - \*R\*{\bar{B}}_i\*c^\star(\*R)\right\|^2 + \lambda\|\*c^\star(\*R)\|^2.
    \end{equation}
    \vspace{3pt}
\end{problem}

Note that \eqref{eq:paceqcqp_rotonly} is an optimization problem over only a single rotation with no constraints beyond $\*R\in\SO3$. Although it is not immediately clear from the formulation, it is also \emph{quadratic} in the unknown matrix $\*R$. In the next section we rewrite~\eqref{eq:paceqcqp_rotonly} as a quartic problem for the quaternion rotation representation and derive its first-order optimality conditions.

\subsection{First-Order Conditions in Terms of Quaternions}
Here we depart from~\cite{Shi23tro-PACE} to derive a fast local solver. Instead of vectoring the rotation matrix $\*R$, we rewrite the problem in terms of the unit quaternion rotation representation. The quaternion formulation has a quartic objective and a quadratic equality constraint, leading to a nonlinear eigenproblem for first-order stationary points.

We begin by expanding the objective of~\eqref{eq:paceqcqp_rotonly}, grouping terms by their dependency on $\*R$. Defining an auxillary variable $\*s \triangleq \sum_{j=1}^N\*{\bar{B}}_j^\T \*R^\T\*{\bar y}_j$, we obtain the quadratic:
\begin{align}
    \min_{\*R\in\SO3} 
&\notag
        \sum_{i=1}^N(\*{\bar{y}}_i^\T\*{\bar{y}}_i) + \*c_2^\T\*{\hat{B}}^2\*c_2 + \lambda\*c_2^\T\*c_2
    \\
    &+
2\*s^T
    \left(
        - \eye_3
        + \*C_1\*{\hat{B}}^2
        + \lambda\*C_1
    \right)
    \*c_2
    \label{eq:paceqcqp_expanded}
    \\
    &+\notag
\*s^\T
    \left(
        - 2\eye_3
        + \*C_1\*{\hat{B}}^2
        + \lambda \*C_1
    \right)
    \*C_1
    \*s.
\end{align}

Now, we rewrite~\eqref{eq:paceqcqp_expanded} in terms of a unit quaternion $\*q\in\mathbb{S}^3$ which represents the same orientation as $\*R$. Using~\prettyref{lem:quatidentity} and dropping the constant terms in~\eqref{eq:paceqcqp_expanded}, we arrive at the following optimization problem:

\begin{equation}
    \label{eq:qcqp_quat}
    \min_{\*q\in\mathbb{S}^3}
    \*q^\T(2\*D + \*A(\*q\*q^\T))\*q,
\end{equation}
where
\begin{equation}
    \label{eq:Cdelta}
    \*C_\delta \triangleq
        \eye_3
        - \*C_1\*{\hat{B}}^2
        - \lambda\*C_1,
\end{equation}
\begin{equation}
    \label{eq:D}
    \*D \triangleq 
    \sum_{i=1}^N
    \*\Omega_l(\*{\bar{y}}_i)
    \*\Omega_r\!\!\left(
    \*{\bar{B}}_i
    \*C_\delta
    \*c_2
    \right)\!,
\end{equation}
\begin{equation}
    \label{eq:A}
    \*A(\*q\*q^\T) \triangleq
    \sum_{i=1}^N
    \*\Omega_l(\*{\bar y}_i)
    \*\Omega_r\!\!\left[
    \*{\bar B}_i
    \left(
        \eye_3 + \*C_\delta
    \right)
    \*C_1
    \*s(\*q)
    \right]\!.\!\!
\end{equation}

The matrix $\*D$ rewrites the linear term in~\eqref{eq:paceqcqp_expanded} into a quadratic quaternion form using~\prettyref{lem:quatidentity} and absorbs the negative into $\*C_\delta$. In contrast, $\*A(\*q\*q^\T)$ uses~\prettyref{lem:quatidentity} on only the first rotation. We write $\*A(\*q\*q^\T)$ to emphasize its remaining quadratic dependence on $\*q$.

We now write the first-order optimality conditions for~\eqref{eq:qcqp_quat}, which are necessary but not sufficient for globally optimal solutions. Introducing the dual variable $\mu$ for the constraint $\*q\in\mathbb{S}^3$, the Lagrangian is:
\begin{equation}
    \label{eq:lagrangian}
    \mathcal{L}(\*q, \mu) = \*q^\T(2\*D + \*A(\*q\*q^\T))\*q + \mu(\*q^\T \*q - 1).
\end{equation}

The first-order conditions are $\nabla_{\*q} \mathcal{L} = \*0$. We differentiate each term with the product rule, noting $\*D$ and $\*A(\*q\*q^\T)$ are symmetric. The $\*q^\T\*A(\*q\*q^\T)\*q$ term requires some care. Numbering each quaternion,~\eqref{eq:A} gives:
\begin{equation}
    \*q^\T_1\*A(\*q_2\*q_3^\T)\*q_4 = \*q^\T_2\*A(\*q_1\*q_4^\T)\*q_3.
\end{equation}

Thus, the derivative of $\*q^\T\*A(\*q\*q^\T)\*q$ term yields four copies. The first-order conditions for~\eqref{eq:qcqp_quat} are:
\begin{equation}
    \label{eq:messystationarity}
    \*0 = \nabla_{\*q} \mathcal{L}(\*q,\mu) =
    4\*A(\*q\*q^\T)\*q + 4\*D\*q - 2\mu\*q.
\end{equation}

Eq. \eqref{eq:messystationarity} resembles an eigenvalue problem, with one summand $\*A(\*q\*q^\T)$ having eigenvector dependence.
\begin{proposition}[Eigenproblem for Local Solutions]
    \label{prop:eigenprob}
    All local minima\footnote{Notice that both $+\*q$ and $-\*q$ are valid eigenvectors, consistent with the double coverage property of quaternions.}
    $\*q$ of \eqref{eq:paceqcqp_rotonly} satisfy the following nonlinear eigenproblem for some $\mu\in\RR$:
    \begin{equation}
        \label{eq:nepv}
        \left(\*A(\*q\*q^\T) + \*D\right)\*q = \mu\*q.
    \end{equation}
\end{proposition}

While it is not immediately clear how to solve~\eqref{eq:nepv}, the weak dependence on $\*q$ suggests standard numerical eigenvalue solvers such as power iteration~\cite{VonMises29-powerIteration} may yield good solutions. Similar eigenproblems have been studied in math and physics~\cite{Martin04-electronicStructure,Cai18-nepv,Zhang20-eigenvalueProcrustes,Li24-NepvStiefel}, but have found less applications in robotics so far. In~\prettyref{sec:scf}, we apply these results to develop a fast iterative solver that only requires computing $\*A(\*q\*q^\T)\in\RR^4$ and its smallest eigenvalue-eigenvector pair at each iteration.  \section{Iterative Method for Fast Shape and Pose Estimates}
\label{sec:scf}
\begin{figure}[tb!]
    \centerline{\includegraphics[width=\linewidth]{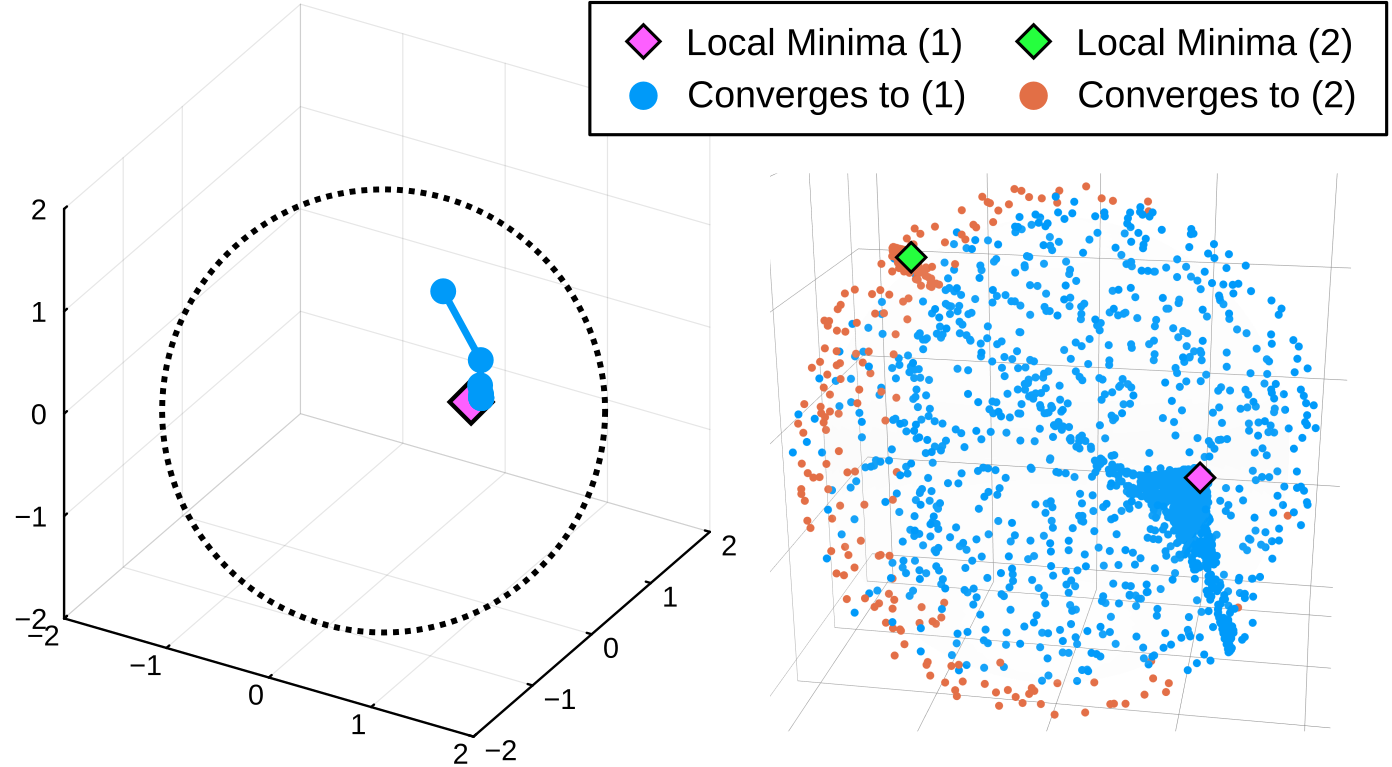}}
    \caption{\textbf{Stereographic projections of self-consistent field iterates.} Beginning from a unit quaternion $\*q_0\in\mathbb{S}^3$, SCF rapidly converges to a local stationary point. Left, a single SCF trajectory.
    Right, unit quaternion iterates stereographically projected into the volume of the $3$-dimensional unit ball 
(see~\prettyref{appendix:scf_proj})
    and colored by which of the two local minima SCF converges to. Nearby starting points tend to converge to the same local minimum except at the distinct boundary. Plots show synthetic data with high measurement noise ($\sigma_m = 5$).
    \vspace{-10pt}
    }
    \label{fig:scf_visual}
\end{figure}

In this section, we propose a fast solution strategy for~\eqref{eq:nepv} using self-consistent field iteration~\cite{Cai18-nepv,Zhang20-eigenvalueProcrustes,Li24-NepvStiefel}. At each iterate, the dominant computational cost is computing the smallest eigenvector-eigenvalue pair of a $4\times4$ matrix. While the solutions are only guaranteed to be local stationary points, we also present a fast certificate of global optimality based on Shor's relaxation~\cite{Shor87}.

\subsection{Self-Consistent Field Iteration}
We use self-consistent field (SCF) iteration to solve the nonlinear eigenproblem~\eqref{eq:nepv}. Starting from an initial guess, SCF computes the data matrix $\*A(\*q\*q^\T) + \*D$ and updates $\*q$ to one of its normalized eigenvectors. The algorithm terminates when it converges to a stationary point---a unit vector $\*q$ which exactly satisfies \eqref{eq:nepv}. In practice, we terminate by a numerical tolerance on the angle between consecutive iterates.

\begin{algorithm}[t]
    \DontPrintSemicolon
    \KwData{$\*A(\*q\*q^\T)$ from~\eqref{eq:A} and $\*D$ from \eqref{eq:D}}
    \KwResult{$\*q$ satisfying \eqref{eq:nepv}}
    initialize $\*q_0\in\mathbb{S}^3$\;
    \For{$t \leftarrow 1$ \KwTo $T$}{
        $\*q_{t+1} \leftarrow \argmin_{\*q \in \mathbb{S}^3} \*q^\T(\*A(\*q_t) + \*D)\*q$\;
        \tcc{termination condition}
        \If{$\sin\measuredangle(\*q_{t},\*q_{t+1}) < \eps$}{
            $\*q\leftarrow\*q_t$\;
            break\;
        }
    }
    \caption{SCF iteration for local solutions to~\eqref{eq:qcqp_quat}.\vspace{-15pt}}
    \label{alg:scf}
\end{algorithm}

The full algorithm is given in \prettyref{alg:scf} and illustrated in \prettyref{fig:scf_visual}. At each iterate, we update $\*q$ according to the eigenvector corresponding to the \emph{minimum} eigenvalue. Although we could pick any of the eigenvectors, picking the smallest has several desirable properties. First, it is likely to be a local minima (rather than a saddle point or local maxima) since the objective at stationary points is dominated by the eigenvalue. To see this, notice that the objective of \eqref{eq:qcqp_quat} can be written as $\*q^\T\left(\*A(\*q\*q^\T) + \*D\right)\*q + \*q^\T\*D\*q$. At a stationary point, \eqref{eq:nepv} implies the first term is just the eigenvalue. That is, $f_\text{local} = \mu + \*q^\T\*D\*q$. Thus, the minimum eigenvalue is a good guess for the local minimum. It also has strong computational benefits. In particular, we observe that the minimum eigenvalue (guaranteed to be non-positive) is often the eigenvalue with largest magnitude at optimality. This property enables fast convergence similar to power iteration~\cite{VonMises29-powerIteration}, often in less than $5$ iterations.

The key advantage of our approach is its speed. A single iteration of SCF requires only computing a $4\times4$ matrix and its minimum eigenvector. The termination condition requires only checking the value of an inner product. For $10$ keypoints, these steps take less than $10\ \mathrm{\mu s}$ on a single CPU thread. Although we do not do so here, starting with different initial conditions could be easily parallelized across GPU resources. In \prettyref{sec:experiments} we show the entire algorithm takes about $100\ \mathrm{\mu s}$.

\subsection{SDP KKT Conditions for Global Optimality}
While the previous sections focused on local solutions to~\prettyref{prob:rotest}, we now develop a tool to \emph{certify} the global optimality of a local solution. Certification is essential for reliability; a certificate guarantees \prettyref{alg:scf} converged to a statistically optimal estimate.

\begin{table}[tb]
    \centering
    \caption{KKT optimality conditions for the QCQP~\eqref{eq:paceqcqp_standard} and its SDP relaxation~\eqref{eq:sdp}.\vspace{-15pt}}
    \label{tab:kkt}
    \adjustbox{width=\linewidth}
    {\begin{tabular}{rcc}
        \toprule
        & QCQP~\eqref{eq:paceqcqp_standard} & SDP~\eqref{eq:sdp}\\
        \midrule
        Stationarity & $\*S\*x = \*0$ & $\*S = \*C - \sum_{i=1}^7 \lambda_i \*A_i$\\
        Slackness &  & $\langle\*S, \*X\rangle = 0$\\
        Primal feas. & $\langle \*A_i, \*x\*x^\T \rangle = b_i$
        & $\langle \*A_i, \*X \rangle = b_i$\\
        Dual feas. & & $\*S \succeq 0$\\
        \bottomrule
    \end{tabular}}
\end{table}

To certify a local solution of~\eqref{eq:qcqp_quat}, we check if it is a KKT point of the problem's convex semidefinite program (SDP) relaxation. To avoid a second-order relaxation, we use the rotation matrix form \eqref{eq:paceqcqp_rotonly}. Our certificate relies on dual variables, but the matrix form of the $\SO3$ constraint does not satisfy the linear independence constraint qualification (necessary for unique duals~\cite{Papalia24tro-coraSLAM}). Thus, we relax~\eqref{eq:paceqcqp_rotonly} to an \emph{orthogonal} matrix constraint~\cite{Boumal23-introManifolds}:
\begin{equation}
    \label{eq:paceqcqp_orthogonal}
    \min_{\*R\in\mathrm{O}(3)} \sum_{i=1}^N\left\|\*{\bar{y}}_i - \*R\*{\bar{B}}_i\*c^\star(\*R)\right\|^2 + \lambda\|\*c^\star(\*R)\|^2.
\end{equation}

Noting that~\eqref{eq:paceqcqp_orthogonal} is \emph{quadratic} in $\*R$, we define the homogeneous variable $\*x\triangleq[1, \mathrm{vec}(\*R)^\T]^\T$. The $\mathrm{vec}(\cdot)$ operator stacks the columns of its argument into a vector. In standard form, \eqref{eq:paceqcqp_orthogonal} is:
\begin{equation}
    \label{eq:paceqcqp_standard}
    \min_{\*x\in\mathbb{R}^{10}} \*x^\T \*C \*x\quad\mathrm{s.t.}\ \*x^\T\*A_i\*x = b_i,\ i=1,...,7,
\end{equation}
for the constraints $x_1=1$ and orthogonality, $\*R^\T\*R = \eye_3$.
Exact expressions for $\*C$, $\*A_i$, and $b_i$ are provided in \prettyref{appendix:certificate}.

Now, we further relax~\eqref{eq:paceqcqp_standard} to an SDP using Shor's relaxation~\cite{Shor87}. Observe $\*x^\T\*C\*x = \langle\*C,\*x\*x^\T\rangle$. Relaxing $\*x\*x^\T$ to a matrix $\*X\in\mathbb{S}^{10}$, we arrive at an SDP: 
\begin{equation}
    \label{eq:sdp}
    \min_{\*X\succeq 0} \langle\*C, \*X\rangle\quad\mathrm{s.t.}\ \langle\*A_i,\*X\rangle = b_i,\ i=1,...,7.
\end{equation}

The KKT conditions for~\eqref{eq:paceqcqp_standard} and \eqref{eq:sdp} are given in \prettyref{tab:kkt}~\cite{Papalia24-bmOverview}. For a local solution $\*x_l$ satisfying the QCQP KKT conditions, the candidate SDP solution $\*X = \*x_l \*x_l^\T$ satisfies primal feasibility and complementarity slackness from QCQP stationarity. Thus, we use the dual feasibility condition as a global optimality certificate. To compute $\*S$, we only need the Lagrange multipliers $\'\lambda$. From QCQP stationarity, this is a linear system:
\begin{equation}
    \label{eq:linearsystem}
    \sum_{i=1}^7 \lambda_i \*A_i\*x  = \*C\*x.
\end{equation}

Thus, to certify optimality of a local solution $\*x$ we solve the linear system \eqref{eq:linearsystem} for $\'\lambda$ and check if $\*S \succeq 0$. 
\section{Experiments}
\label{sec:experiments}
\begin{figure}[tb]
    \centerline{\includegraphics[width=\linewidth]{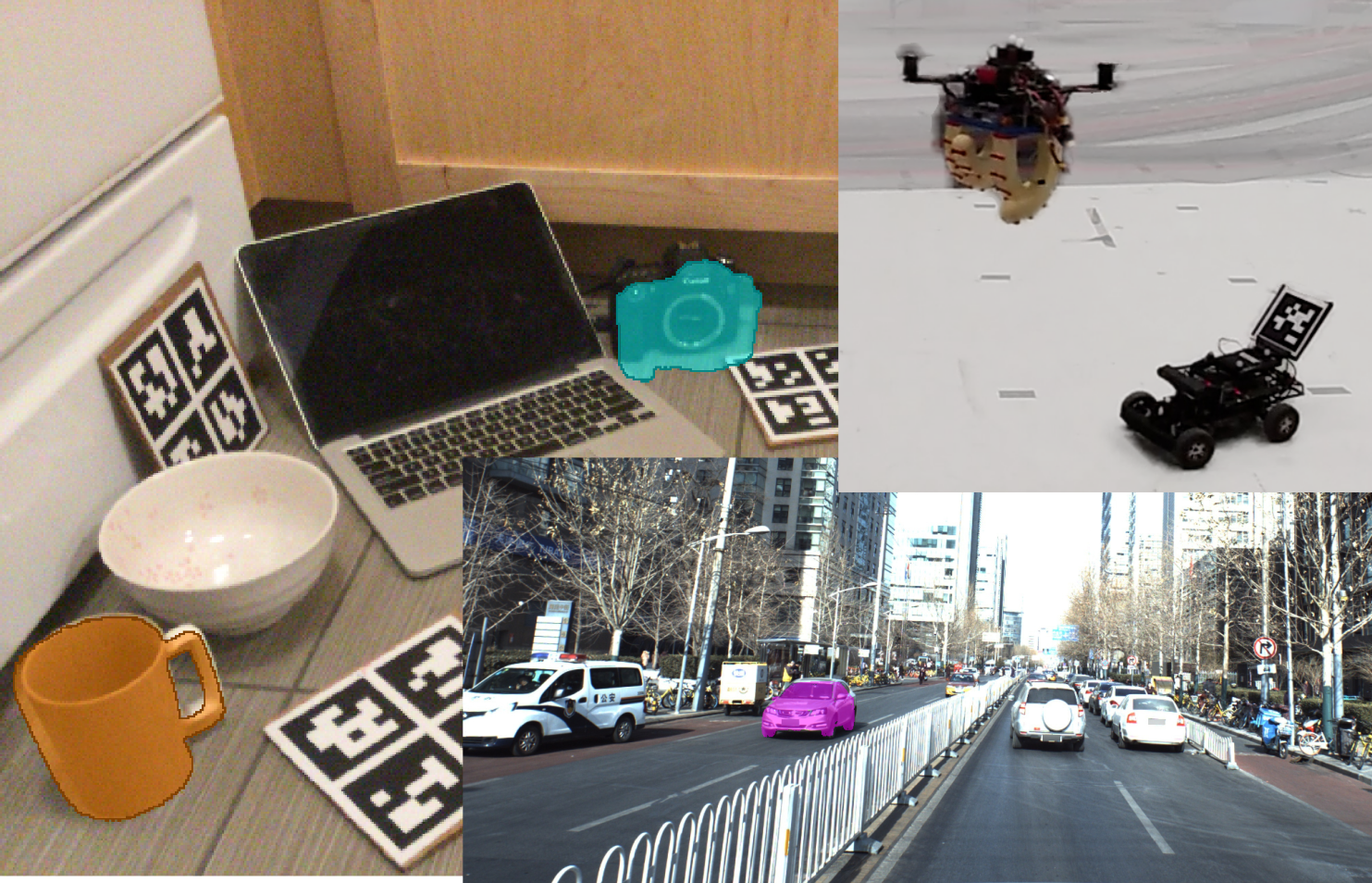}}
    \caption{\textbf{Overview of experiments.}  We test on a variety of datasets and synthetic data (not pictured). Left, NOCS-REAL275~\cite{Wang19-normalizedCoordinate} contains common household categories including mugs and cameras. Upper right, the CAST drone dataset~\cite{Shaikewitz24ral-CAST} includes pictures from an aerial quadcopter following a small racecar. Lower right, ApolloCar3D~\cite{Song19-apollocar3d} has real-world autonomous driving. Sample pose estimates are highlighted in color.
    \vspace{-5pt}
}
    \label{fig:experiments}
\end{figure}

In this section, we evaluate the computational speed and estimation accuracy of our approach against other category-level shape and pose estimators. In~\prettyref{sec:synthetic} we demonstrate that self-consistent field iteration (SCF) produces good local solutions when our Gaussian noise priors are satisfied (\ie no outliers). Next,~\prettyref{sec:cast} extends this evaluation to the real-world drone tracking scenario in~\cite{Shaikewitz24ral-CAST}, where we use graduated non-convexity for outlier robustness~\cite{Yang20ral-GNC}. Lastly, we compare against learning-based methods on the NOCS-REAL275 dataset~\cite{Wang19-normalizedCoordinate} (\prettyref{sec:nocs}) and the ApolloCar3D dataset~\cite{Song19-apollocar3d}. These experiments are summarized in~\prettyref{fig:experiments}. All benchmarks run on a single CPU thread with clock speed $4.2\ \mathrm{GHz}$.

\subsection{Empirical Performance in Synthetic Dataset}
\label{sec:synthetic}
We begin by evaluating SCF in a synthetic environment, where measurements are generated according to the generative model~\eqref{eq:pacemeas} with known Gaussian noise. Specifically, we generate a mean shape with $N=10$ points drawn from a standard normal. The shape library $\*B$ adds zero-mean Gaussian noise to each point with fixed standard deviation $r=0.2\ \mathrm{m}$. We generate the ground truth shape by normalizing a $K=4$ dimensional vector uniformly random in $[0,1]^K$. The ground truth position is drawn from a standard normal with mean $1$ and ground truth rotation is randomly sampled from $\SO3$. We set $\lambda=0$. Results are reported in terms of a normalized measurement standard deviation $\sigma_m \triangleq r/w$, with $w \triangleq w_1 = \hdots = w_N$ the same for all keypoints.

\textbf{Baselines.}
We compare against other solution strategies for~\eqref{eq:paceqcqp_rotonly}. Manopt~\cite{manopt} is an off-the-shelf local solver for unconstrained manifold optimization. Gauss--Newton (G-N)~\cite{Gauss1809} and Levenberg--Marquardt (L-M)~\cite{Levenberg44qam} are local solvers with optimized implementations and analytic Jacobians for~\eqref{eq:paceqcqp_rotonly} (details in \prettyref{appendix:scf_gn}). Lastly, \pace~\cite{Shi23tro-PACE} uses a semidefinite relaxation to find and certify global solutions to~\eqref{eq:paceqcqp_rotonly}. We denote by \scfstar~our approach (SCF) with optimality certificate checking. All methods are implemented in Julia and runtimes do not include precompilation time. For each measurement noise value, each method solves the same 10,000 problems with the same initial guess.

\textbf{Computational Speed.}
\prettyref{tab:scf_runtimes1} compares the mean and 90th percentile (p90) of runtimes for each method at small and large noise scales. SCF is by far the fastest method, and more than twice as fast as G-N or L-M. Certificate checking for \scfstar~adds only a small computational penalty.

\begin{table}[tb]
    \centering
    \caption{Runtimes on synthetic dataset.\vspace{-10pt}}
    \label{tab:scf_runtimes1}
    \adjustbox{width=\linewidth}
    {\begin{tabular}{rcccc}
        \toprule
        &      \multicolumn{2}{c}{$\sigma_m =0.25$} & \multicolumn{2}{c}{$\sigma_m = 2.5$} \\
        \cmidrule(lr){2-3}\cmidrule(lr){4-5}
        &      Mean (ms)   & p90 (ms)   & Mean (ms)  & p90 (ms)  \\ \midrule
        SCF &        \textbf{0.104} &      \textbf{0.117} &      \textbf{0.108} &     \textbf{0.126} \\ 
        G-N &        0.217 &      0.253 &      0.299 &     0.366 \\ 
        L-M &        0.211 &      0.244 &      0.285 &     0.345 \\ 
        Manopt &       1.028 &      1.155 &      1.039 &     1.164 \\ \midrule
        \scfstar &        0.120 &      0.126 &      0.126 &     0.138 \\ 
        \pace &       1.655 &      1.601 &      1.667 &     1.617 \\
        \bottomrule
    \end{tabular}}
    \vspace{-5pt}
\end{table}

\begin{figure}[tb]
    \centerline{\includegraphics[width=\linewidth]{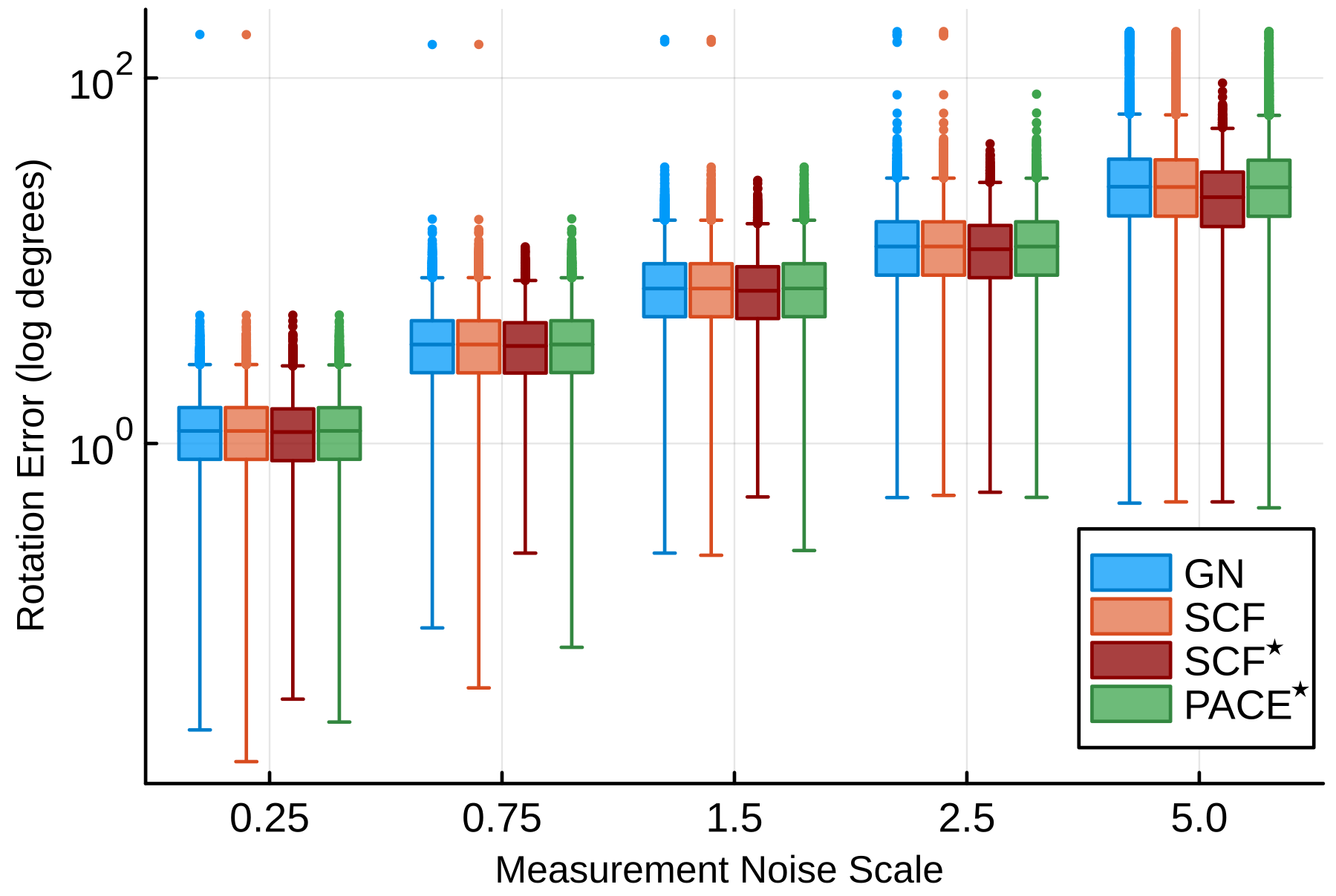}}
    \caption{\textbf{Distribution of rotation errors for Gauss-Newton, SCF, \scfstar, and \pace.} 
    GN, SCF, and PACE have nearly identical performance although SCF runs significantly faster. \scfstar~and \pace~show only certifiably optimal estimates. \scfstar~consistently filters out the worst estimates.
    \vspace{-15pt}
    }
    \label{fig:scf_mnoise}
\end{figure}

\textbf{Estimation Performance.}
We also compare the rotation error distributions of G-N, SCF, \scfstar, and \pace~in \prettyref{fig:scf_mnoise} (we omit the other methods for clarity). G-N, SCF, and \pace~exhibit similar performance across noise scales, with the local methods (G-N and SCF) having slightly more outliers throughout. In view of~\prettyref{tab:scf_runtimes1}, we conclude SCF is much faster without sacrificing accuracy. For this plot only, \scfstar~and \pace~results show only \emph{globally optimal} estimates. For \pace, all estimates reported a global optimality certificate (using tolerance $10^{-4}$). \scfstar~certified global optimality for $62$\%, $60$\%, $55$\%, $45$\%, and $19$\% of estimates at noise scales $0.25$, $0.75$, $1.5$, $2.5$, and $5.0$, respectively. \scfstar~reports fewer certificates because of the relaxation from $\SO3$ to $\mathrm{O}(3)$, but the certified estimates are, on average, more accurate.

\subsection{Performance in Drone Tracking Scenario}
\label{sec:cast}
We also test SCF in the real-world drone tracking scenario from~\cite{Shaikewitz24ral-CAST} (CAST). In the CAST dataset, an autonomous quadcopter follows a remote-controlled racecar around a track. There are $1897$ RGB-D images taken in sequence from the quadcopter. For this and following experiments, real-world keypoint measurements are corrupted by outliers. Thus, we first run compatibility tests~\cite{Shi20tr-robin} and wrap SCF in graduated non-convexity (GNC)~\cite{Yang20ral-GNC}. Both use $\*{\bar c}^2 = 0.005$. We compare with the same baselines as~\prettyref{sec:synthetic} and use identical outlier rejection. We use the resnet-based~\cite{He16cvpr-ResNet} keypoint detections ($N=7$) and shape library ($K=10$) from~\cite{Shaikewitz24ral-CAST} with $\lambda=0.1$.

\textbf{Results.}
We report total runtime (which includes GNC and compatibility tests but not keypoint detection), mean rotation error ($R_\mathrm{err}$), and mean GNC iterations (number of times the solver is called) in \prettyref{tab:cast_results}. Recall that~\prettyref{prob:rotest} is a rotation estimation problem, so rotation error is a good proxy for overall estimation error. As before, SCF is substantially faster than other local approaches and nearly five times faster than G-N. The runtime per iteration is faster than synthetic experiments because of fewer keypoints ($N=7$). All methods achieve very similar rotation estimation performance and require the same number of GNC iterations. 

\begin{table}[tb]
    \centering
    \caption{Performance on CAST drone dataset~\cite{Shaikewitz24ral-CAST}.
    \vspace{-15pt}}
    \label{tab:cast_results}
    \adjustbox{width=\linewidth}
    {\begin{tabular}{rcccc}
    \toprule
    &      \multicolumn{2}{c}{Runtimes} & $R_\mathrm{err}$ & GNC Iters.\\
        \cmidrule(lr){2-3}
                              &  Mean (ms)   & p90 (ms) & Mean (deg) & Mean   \\ \midrule
    \multirow{1}{*}{SCF}      &  \textbf{0.456} & \textbf{0.774} & 9.4 & 7.1\\
        \multirow{1}{*}{G-N}  &  1.817 &  3.480 & 9.5 &  7.1\\
        \multirow{1}{*}{L-M}  &  5.200 & 12.701 & 9.5 &  7.1\\
    \multirow{1}{*}{Manopt}   &  6.062 & 11.783 & 9.5 &  7.1\\\midrule
    \multirow{1}{*}{\scfstar} &  0.606 &  1.029 & 9.4 &  7.1\\
    \multirow{1}{*}{\pace}    & 10.826 & 21.514 & 9.5 &  7.1\\
    \bottomrule
    \end{tabular}}
\end{table}

\subsection{Performance on NOCS-REAL275 Dataset}
\label{sec:nocs}
The NOCS-REAL275 dataset~\cite{Wang19-normalizedCoordinate} contains real-world RGB-D video sequences of common objects within $6$ categories. We test on the camera (2561 frames, $N=50$, $K=3$) and mug (2615 frames, $N=43$, $K=5$) categories. We drop the other objects due to keypoint detector availability. We use the same YOLOv8~\cite{ultralytics23_yolov8} detector as~\cite{Shaikewitz24ral-CAST}, which was trained using synthetically-generated images. We also use the same shape library, which is composed of representative CAD models from 3D scans and BOP datasets~\cite{Hodan20eccvw-BOPChallenge} (note that these are distinct from the objects in NOCS). We compare against the same set of baselines and the tracking method BundleTrack~\cite{Wen21iros-bundletrack}. BundleTrack numbers are from~\cite{Wen21iros-bundletrack}.

\textbf{Results.} In~\prettyref{tab:nocs_results}, we report the mean runtime across categories and accuracy for each category. For accuracy, we report the percentage of estimates within $5^\circ$ and $5\ \mathrm{cm}$ of ground truth ($5^\circ5\mathrm{cm}$) and the mean orientation error in degrees ($R_\mathrm{err}$) and position error in cm ($p_\mathrm{err}$). For position and orientation error we exclude measurements with position error above $10\ \mathrm{cm}$~\cite{Wen21iros-bundletrack}. We exclude keypoint detection runtime. For reference, the keypoint detector runs in about $50\ \mathrm{ms}$ per image. SCF is significantly faster than other methods, running in just over a millisecond (the majority of this runtime is from compatibility tests). The two-stage methods achieve similar performance, although they are significantly worse than BundleTrack except for mug position error. This poor performance is largely due to the low-quality keypoint detector~\cite{Shaikewitz24ral-CAST}.

\begin{table}[tb]
    \centering
    \caption{NOCS-REAL275~\cite{Wang19-normalizedCoordinate} Performance.}
    \label{tab:nocs_results}
    \setlength{\tabcolsep}{3pt}
    \adjustbox{width=\linewidth}
    {\begin{tabular}{rcccccccccc}
    \toprule
    & \multicolumn{3}{c}{camera} & \multicolumn{3}{c}{mug} & Time\\
        \cmidrule(lr){2-4} \cmidrule(lr){5-7}
    & $5^\circ5\mathrm{cm}$ & $R_\mathrm{err}$ & $p_\mathrm{err}$
    & $5^\circ5\mathrm{cm}$ & $R_\mathrm{err}$ & $p_\mathrm{err}$
    & mean (ms)
    \\\midrule
    SCF         
                & 7.8 & 19.5 & 3.4
                & 24.0 & 12.7 & \textbf{1.0} & \textbf{1.26}\\
    G-N         
                & 7.8 & 18.4 & 3.4
                & 23.6 & 12.8 & 1.1 & 1.81\\
    L-M         
                & 6.5 & 21.6 & 3.4
                & 21.6 & 12.5 & \textbf{1.0} & 49.1\\
    Manopt      
                & 7.8 & 18.2 & 3.4
                & 23.5 & 12.5 & \textbf{1.0} & 2.45\\
    BundleTrack          
                & \textbf{85.8} & \textbf{3.0} & \textbf{2.1}
                & \textbf{99.9} & \textbf{1.5} & 2.2 & 100\\
    \midrule
    \scfstar    
                & 7.8 & 18.9 & 3.4
                & 23.4 & 12.9 & \textbf{1.0} & 1.34\\
    \pace       
                & 7.7 & 15.0 & 3.4
                & 20.2 & 11.9 & \textbf{1.0} & 3.90\\
    \bottomrule
    \end{tabular}}
\end{table}

\begin{table}[tb]
    \centering
    \caption{ApolloCar3D~\cite{Wang19-normalizedCoordinate} Performance.\vspace{-10pt}}
    \label{tab:apollo_results}
\adjustbox{width=\linewidth}
    {\begin{tabular}{rccccc}
    \toprule
    & \multicolumn{3}{c}{A3DP-Rel $\uparrow$} & GNC Iters. & Time\\
        \cmidrule(lr){2-4}
    & mean & c-l & c-s
    & mean
    & mean (ms)
    \\\midrule
    SCF         & 17.1 & 35.7 & \textbf{28.4} & 5.8 & \textbf{4.4}\\
    G-N         & 17.1 & 35.5 & \textbf{28.4} & 5.6 & 10.3\\
    L-M         & 17.1 & 35.7 & 28.3 & 5.4 & 8.0\\
    Manopt      & 17.1 & 35.7 & \textbf{28.4} & 6.9 & 17.7\\
    GSNet       & \textbf{20.2} & \textbf{40.5} & 19.9 & - & 450\\
    \midrule
    \scfstar    & 17.1 & 35.7 & \textbf{28.4} & 5.8 & 4.8\\
    \pace       & 17.2 & 35.7 & \textbf{28.4} & 5.3 & 11.2\\
    \bottomrule
    \end{tabular}}
\end{table}

\subsection{Performance on ApolloCar3D}
\label{sec:cars}
Lastly, we evaluate SCF on the autonomous driving dataset ApolloCar3D~\cite{Song19-apollocar3d}. This dataset has real-world stereo images taken from cars driving in four cities in China. We test on the $200$-image validation split and use $K=79$ car models with $N=66$ semantic keypoint annotations. For this larger shape library, we set $K=1.5\cdot10^{4}$. As in~\cite{Shi23tro-PACE}, we use the keypoint detections from GSNet~\cite{Ke20-gsnet} with stereo depths. For GNC and compatibility tests, we set $\*{\bar c}^2 = 0.15\ \mathrm{m}$. We compare with GSNet~\cite{Ke20-gsnet} and the same set of solvers as in synthetic experiments.

\textbf{Results.} The estimation performance and runtimes are shown in~\prettyref{tab:apollo_results}. For estimation, we use the A3DP-Rel metric as defined in~\cite{Song19-apollocar3d}. This metric jointly measures translation, rotation, and shape similarity between the estimated and ground truth cars using relative translation thresholds. ``Mean'' averages across $10$ thresholds, while ``c-l'' reports performance under a loose threshold and ``c-s'' reports performance under a strict threshold. We also report the mean number of GNC iterations and the mean runtime of each approach. We exclude keypoint detector time and use GSNet results from~\cite{Ke20-gsnet}. SCF is conclusively faster than the other approaches, and significantly outperforms GSNet on the strict (c-s) criterion. It is not far behind in the mean and loose criteria. The GSNet runtime is difficult to compare with because it includes keypoint detection and was evaluated on different hardware. Although we attempted to exactly replicate~\cite{Shi23tro-PACE}, our accuracy results are slightly worse.
 \section{Conclusion}
In this paper, we revisited the shape and pose estimation problem in~\cite{Shi23tro-PACE} with an emphasis on solver speed. Our local solver, based on self-consistent field iteration, can estimate an object's shape and pose in about $100$ microseconds. The solver exploits the structure of the quaternion form of the first-order optimality conditions for~\prettyref{prob:rotest}, which are a nonlinear eigenvalue problem for stationary points. We also augmented this local solver with a fast global optimality certificate based on duality with the SDP relaxation. In synthetic and real-world experiments, we demonstrated that our solver was significantly faster than other approaches, with and without global optimality checks. The mixed estimation results suggest more work is needed to develop fast and accurate semantic keypoint detectors.

\bibliographystyle{plain}
\scriptsize{
    % \bibliography{biber.bib}

\begin{thebibliography}{}
        \expandafter\ifx\csname natexlab\endcsname\relax\def\natexlab#1{#1}\fi
        \def\au#1{#1} \def\ed#1{#1} \def\yr#1{#1}\def\at#1{#1}\def\jt#1{{#1}}
        \def\bt#1{#1}\def\bvol#1{\textbf{#1}} \def\vol#1{#1} \def\pg#1{#1}
        \def\publ#1{#1}\def\arxiv#1{#1}\def\org#1{#1}\def\st#1{\textit{#1}}\def\series#1{#1}
    \bibitem{Koopman16sae-autonomousVehicleTesting}
{\au{P. Koopman} and \au{M. Wagner}}, \at{\textit{Challenges in Autonomous Vehicle Testing and Validation}}, \jt{SAE Int. J. Trans. Safety}, \bvol{4}(1) (\yr{2016}).

\bibitem{Tremblay18corl-DeepObject}
{\au{J. Tremblay}, \au{T. To}, \au{B. Sundaralingam}, \au{Y. Xiang}, \au{D. Fox}, and \au{S. Birchfield}}, \at{\textit{Deep Object Pose Estimation for Semantic Robotic Grasping of Household Objects}}, \bt{Conference on Robot Learning (CoRL)}, (\yr{2018}), \pg{{306}--{316}}.

\bibitem{Shi23tro-PACE}
{\au{J. Shi}, \au{H. Yang}, and \au{L. Carlone}}, \at{\textit{Optimal and Robust Category-level Perception: Object Pose and Shape Estimation from 2D and 3D Semantic Keypoints}}, \jt{IEEE Trans. Robotics}, \bvol{39}(5) (\yr{2023}) \pg{{4131}--{4151}}.

\bibitem{Yang20ral-GNC}
{\au{H. Yang}, \au{P. Antonante}, \au{V. Tzoumas}, and \au{L. Carlone}}, \at{\textit{Graduated Non-Convexity for Robust Spatial Perception: From Non-Minimal Solvers to Global Outlier Rejection}}, \jt{IEEE Robotics and Automation Letters (RA-L)}, \bvol{5}(2) (\yr{2020}) \pg{{1127}--{1134}}.

\bibitem{Fischler81}
{\au{M. Fischler} and \au{R. Bolles}}, \at{\textit{Random sample consensus: a paradigm for model fitting with application to image analysis and automated cartography}}, \jt{Commun. ACM}, \bvol{24} (\yr{1981}) \pg{{381}--{395}}.

\bibitem{Cai18-nepv}
{\au{Y. Cai}, \au{L-H. Zhang}, \au{Z. Bai}, and \au{R-C. Li}}, \at{\textit{On an Eigenvector-Dependent Nonlinear Eigenvalue Problem}}, \jt{SIAM Journal on Matrix Analysis and Applications}, \bvol{39}(3) (\yr{2018}) \pg{{1360}--{1382}}.

\bibitem{Gauss1809}
{\au{C. Gauss}}, \at{\textit{Theoria Motus Corporum Coelestium in Sectionibus Conicis Solem Mabientium [Theory of the Motion of the heavenly Bodies Moving about the Sun in Conic Sections]}},  (\yr{1809}) \publ{Perthes}.

\bibitem{Levenberg44qam}
{\au{K. Levenberg}}, \at{\textit{A method for the solution of certain nonlinear problems in least squares}}, \jt{Quart. Appl. Math}, \bvol{2}(2) (\yr{1944}) \pg{{164}--{168}}.

\bibitem{manopt}
{\au{N. Boumal}, \au{B. Mishra}, \au{P-A. Absil}, and \au{R. Sepulchre}}, \at{\textit{Manopt, a Matlab Toolbox for Optimization on Manifolds}}, \jt{Journal of Machine Learning Research}, \bvol{15}(42) (\yr{2014}) \pg{{1455}--{1459}}.

\bibitem{Hertzberg08-manifolds}
{\au{C. Hertzberg}}, \at{\textit{A framework for sparse, non-linear least squares problems on manifolds}}, \jt{UNIVERSIT AT BREMEN} (\yr{2008}).

\bibitem{Chen22cvpr-manifoldPGD}
{\au{J. Chen}, \au{Y. Yin}, \au{T. Birdal}, \au{B. Chen}, \au{L. J. Guibas}, and \au{H. Wang}}, \at{\textit{Projective manifold gradient layer for deep rotation regression}}, \bt{Proceedings of the IEEE/CVF Conference on Computer Vision and Pattern Recognition}, (\yr{2022}), \pg{{6646}--{6655}}.

\bibitem{Smith94ams}
{\au{S. T. Smith}}, \at{\textit{Optimization techniques on Riemannian manifolds}}, \jt{Hamiltonian and Gradient Flows, Algorithms and Control, Fields Inst. Commun., Amer. Math. Soc.}, \bvol{3} (\yr{1994}) \pg{{113}--{136}}.

\bibitem{Shor87}
{\au{N. Shor}}, \at{\textit{Quadratic optimization problems}}, \jt{Izv. Akad. Nauk SSSR Tekhn. Kibernet.}, \bvol{1} (\yr{1987}) \pg{{128}--{139}}.

\bibitem{Burer03mp}
{\au{S. {Burer, Samuel}} and \au{R. {Monteiro, Renato D C}}}, \at{\textit{A nonlinear programming algorithm for solving semidefinite programs via low-rank factorization}}, \jt{Mathematical Programming}, \bvol{95}(2) (\yr{2003}) \pg{{329}--{357}}.

\bibitem{Brynte21-semidefiniteRotation}
{\au{L. Brynte}, \au{V. Larsson}, \au{J. P. Iglesias}, \au{C. Olsson}, and \au{F. Kahl}}, \at{\textit{On the Tightness of Semidefinite Relaxations for Rotation Estimation}}, \jt{Journal of Mathematical Imaging and Vision}, \bvol{64}(1) (\yr{2021}) \pg{{57}--{67}}.

\bibitem{Rosen18ijrr-sesync}
{\au{D. Rosen}, \au{L. Carlone}, \au{A. Bandeira}, and \au{J. Leonard}}, \at{\textit{SE-Sync: a certifiably correct algorithm for synchronization over the Special Euclidean group}}, \jt{Intl. J. of Robotics Research} (\yr{2018}).

\bibitem{Shaikewitz24ral-CAST}
{\au{L. Shaikewitz}, \au{S. Ubellacker}, and \au{L. Carlone}}, \at{\textit{A Certifiable Algorithm for Simultaneous Shape Estimation and Object Tracking}}, \jt{IEEE Robotics and Automation Letters (RA-L)} (\yr{2024}).

\bibitem{Arun87pami}
{\au{K. Arun}, \au{T. Huang}, and \au{S. Blostein}}, \at{\textit{Least-Squares Fitting of Two 3-D Point Sets}}, \jt{IEEE Trans. Pattern Anal. Machine Intell.}, \bvol{9}(5) (\yr{1987}) \pg{{698}--{700}}.

\bibitem{Shuster89jas}
{\au{M. Shuster}}, \at{\textit{Maximum Likelihood Estimation of Spacecraft Attitude}}, \jt{J. Astronautical Sci.}, \bvol{37}(1) (\yr{1989}) \pg{{79}--{88}}.

\bibitem{Ke20-gsnet}
{\au{L. Ke}, \au{S. Li}, \au{Y. Sun}, \au{Y-W. Tai}, and \au{C-K. Tang}}, \at{\textit{GSNet: Joint Vehicle Pose and Shape Reconstruction with Geometrical and Scene-aware Supervision}}, \bt{European Conf. on Computer Vision (ECCV)}, (\yr{2020}), \pg{{515}--{532}}.

\bibitem{Wen21iros-bundletrack}
{\au{B. Wen} and \au{K. Bekris}}, \at{\textit{Bundletrack: 6d pose tracking for novel objects without instance or category-level 3d models}}, \bt{IEEE/RSJ Intl. Conf. on Intelligent Robots and Systems (IROS)}, (\yr{2021}), \pg{{8067}--{8074}}.

\bibitem{Liu22eccv-CATRE}
{\au{X. Liu}, \au{G. Wang}, \au{Y. Li}, and \au{X. Ji}}, \at{\textit{Catre: Iterative point clouds alignment for category-level object pose refinement}}, \bt{European Conference on Computer Vision}, (\yr{2022}), \pg{{499}--{516}}.

\bibitem{Chen24cvpr-secondpose}
{\au{Y. Chen}, \au{Y. Di}, \au{G. Zhai}, \au{F. Manhardt}, \au{C. Zhang}, \au{R. Zhang}, \au{F. Tombari}, \au{N. Navab}, and \au{B. Busam}}, \at{\textit{Secondpose: Se (3)-consistent dual-stream feature fusion for category-level pose estimation}}, \bt{Proceedings of the IEEE/CVF Conference on Computer Vision and Pattern Recognition}, (\yr{2024}), \pg{{9959}--{9969}}.

\bibitem{Wang19-normalizedCoordinate}
{\au{H. Wang}, \au{S. Sridhar}, \au{J. Huang}, \au{J. Valentin}, \au{S. Song}, and \au{L. Guibas}}, \at{\textit{Normalized object coordinate space for category-level 6d object pose and size estimation}}, \bt{IEEE Conf. on Computer Vision and Pattern Recognition (CVPR)}, (\yr{2019}), \pg{{2642}--{2651}}.

\bibitem{Shi25cvpr-CRISP}
{\au{J. Shi}, \au{R. Talak}, \au{H. Zhang}, \au{D. Jin}, and \au{L. Carlone}}, \at{\textit{CRISP: Object Pose and Shape Estimation with Test-Time Adaptation}}, \bt{IEEE Conf. on Computer Vision and Pattern Recognition (CVPR)}, (\yr{2025}).

\bibitem{Tian20eccv-SPD}
{\au{M. Tian}, \au{M. H. Ang}, and \au{G. H. Lee}}, \at{\textit{Shape prior deformation for categorical 6d object pose and size estimation}}, \bt{European Conf. on Computer Vision (ECCV)}, (\yr{2020}), \pg{{530}--{546}}.

\bibitem{Liu24arxiv-deepLearningSurvey}
{\au{J. Liu}, \au{W. Sun}, \au{H. Yang}, \au{Z. Zeng}, \au{C. Liu}, \au{J. Zheng}, \au{X. Liu}, \au{H. Rahmani}, \au{N. Sebe}, and \au{A. Mian}}, \at{\textit{Deep learning-based object pose estimation: A comprehensive survey}}, \jt{arXiv preprint arXiv:2405.07801} (\yr{2024}).

\bibitem{Altmann13-Rotations}
{\au{S. Altmann}}, \at{\textit{Rotations, Quaternions, and Double Groups}}, \series{Dover Books on Mathematics},  (\yr{2013}) \publ{Dover Publications}.

\bibitem{Yang19iccv-QUASAR}
{\au{H. Yang} and \au{L. Carlone}}, \at{\textit{A Quaternion-based Certifiably Optimal Solution to the Wahba Problem with Outliers}}, \bt{Intl. Conf. on Computer Vision (ICCV)}, (\yr{2019}).

\bibitem{He17iccv-maskRCNN}
{\au{K. He}, \au{G. Gkioxari}, \au{P. Dollár}, and \au{R. Girshick}}, \at{\textit{Mask R-CNN}}, \bt{Intl. Conf. on Computer Vision (ICCV)}, (\yr{2017}), \pg{{2980}--{2988}}.

\bibitem{VonMises29-powerIteration}
{\au{R. V. Mises} and \au{H. Pollaczek-Geiringer}}, \at{\textit{Praktische Verfahren der Gleichungsauflösung.}}, \jt{Journal of Applied Mathematics and Mechanics}, \bvol{9}(2) (\yr{1929}) \pg{{152}--{164}}.

\bibitem{Martin04-electronicStructure}
{\au{R. M. Martin}}, \at{\textit{Electronic Structure: Basic Theory and Practical Methods}},  (\yr{2004}) \publ{Cambridge University Press}.

\bibitem{Zhang20-eigenvalueProcrustes}
{\au{L-H. Zhang}, \au{W. H. Yang}, \au{C. Shen}, and \au{J. Ying}}, \at{\textit{An Eigenvalue-Based Method for the Unbalanced Procrustes Problem}}, \jt{SIAM Journal on Matrix Analysis and Applications}, \bvol{41}(3) (\yr{2020}) \pg{{957}--{983}}.

\bibitem{Li24-NepvStiefel}
{\au{R-C. Li}}, \at{\textit{A Theory of the NEPv Approach for Optimization On the Stiefel Manifold}}, preprint, (\yr{2024}). arXiv:\arxiv{\href{http://arxiv.org/abs/2305.00091}{2305.00091}}.

\bibitem{Papalia24tro-coraSLAM}
{\au{A. Papalia}, \au{A. Fishberg}, \au{B. W. O'Neill}, \au{J. P. How}, \au{D. M. Rosen}, and \au{J. J. Leonard}}, \at{\textit{Certifiably correct range-aided SLAM}}, \jt{IEEE Transactions on Robotics} (\yr{2024}).

\bibitem{Boumal23-introManifolds}
{\au{N. Boumal}}, \at{\textit{An introduction to optimization on smooth manifolds}},  (\yr{2023}) \publ{Cambridge University Press}.

\bibitem{Papalia24-bmOverview}
{\au{A. Papalia}, \au{Y. Tian}, \au{D. M. Rosen}, \au{J. P. How}, and \au{J. J. Leonard}}, \at{\textit{An overview of the Burer-Monteiro method for certifiable robot perception}}, \jt{arXiv preprint arXiv:2410.00117} (\yr{2024}).

\bibitem{Song19-apollocar3d}
{\au{X. Song}, \au{P. Wang}, \au{D. Zhou}, \au{R. Zhu}, \au{C. Guan}, \au{Y. Dai}, \au{H. Su}, \au{H. Li}, and \au{R. Yang}}, \at{\textit{ApolloCar3D: A large 3d car instance understanding benchmark for autonomous driving}}, \bt{IEEE Conf. on Computer Vision and Pattern Recognition (CVPR)}, (\yr{2019}), \pg{{5452}--{5462}}.

\bibitem{Shi20tr-robin}
{\au{J. Shi}, \au{H. Yang}, and \au{L. Carlone}}, \at{\textit{ROBIN: a Graph-Theoretic Approach to Reject Outliers in Robust Estimation using Invariants}}, \jt{arXiv preprint: 2011.03659} (\yr{2020}).

\bibitem{He16cvpr-ResNet}
{\au{K. He}, \au{X. Zhang}, \au{S. Ren}, and \au{J. Sun}}, \at{\textit{Deep residual learning for image recognition}} (\yr{2016}) \pg{{770}--{778}}.

\bibitem{ultralytics23_yolov8}
{\au{G. Jocher}, \au{A. Chaurasia}, and \au{J. Qiu}}, \at{\textit{Ultralytics YOLO}}.

\bibitem{Hodan20eccvw-BOPChallenge}
{\au{T. Hodaň}, \au{M. Sundermeyer}, \au{B. Drost}, \au{Y. Labbé}, \au{E. Brachmann}, \au{F. Michel}, \au{C. Rother}, and \au{J. Matas}}, \at{\textit{BOP Challenge 2020 on 6D Object Localization}}, \jt{European Conference on Computer Vision Workshops (ECCVW)} (\yr{2020}).

\bibitem{Barfoot17book}
{\au{T. Barfoot}}, \at{\textit{State Estimation for Robotics}},  (\yr{2017}) \publ{Cambridge University Press}.
\end{thebibliography}
    
}
\normalsize

\appendices
\renewcommand{\theequation}{A\arabic{equation}}
\renewcommand{\thetheorem}{A\arabic{theorem}}

\section{Stereographic Projection of Unit Quaternions}
\label{appendix:scf_proj}
To visualize unit quaternions in Figs. \ref{fig:scf_visual} and \ref{fig:intro}, we stereographically project them from the $4$-sphere onto the volume of the unit $3$-ball. The projection is simple. Let $\*q\in\mathbb{S}^3$ denote a unit quaternion. Recall that $-\*q$ represents the same rotation. When the scalar part is positive, the vector part can be understood as coordinates within the $3$-ball. When the scalar part is negative, we simply negate the quaternion and use the vector part. Thus, we project by taking the vector part times the sign of the scalar part. For a quaternion $\*q = [q_1, \*q_v^\T]^\T$,
\begin{equation}
    \*q_{\mathrm{proj}} = -\mathrm{sign}(q_1)\*q_v.
\end{equation}

As with any projection this is an imperfect representation of the space. Points which are on the boundary and on opposite sides of the unit ball are actually quite close to each other. The projection also makes the quaternion $\*q$ indistinguishable from its inverse $\*q^{-1} = [q_1, -\*q_v^\T]^\T$.

\section{Exact Objective and Constraint Matrix Expressions}
\label{appendix:certificate}
Here we give the exact expressions for the objective matrix $\*C$ and the constraints $\*A_i$ and $b_i$, $i=1,...,7$, in~\eqref{eq:paceqcqp_standard}. These matrices are mostly zero. We write them in sparse form: $(j,k,l)$ means $(\*A_i)_{j,k} = (\*A_i)_{k,j} = l$.

First, the constraint $i=1$ enforces $x_1 = 1$. Thus, $A_1: (1,1,1)$ and $b_1 = 1$. The remaining constraints enforce $\*R \in \mathrm{O}(3)$. All $b_i = 0$ for $i=2,...,7$. Three matrices enforce the columns of $\*R$ are unit norm:
\begin{equation}
    \begin{array}{rl}
        \*A_2:& (2,2,1), (3,3,1), ( 4,4,1), (1,1,-1).\\
        \*A_3:& (5,2,1), (6,3,1), ( 7,4,1), (1,1,-1).\\
        \*A_4:& (8,2,1), (9,3,1), (10,4,1), (1,1,-1),
    \end{array}
\end{equation}
and three matrices enforce the columns are orthogonal:
\begin{equation}
    \begin{array}{rl}
        \*A_5:& (2,5,1), (3,6,1), (4,7,1).\\
        \*A_6:& (2,8,1), (3,9,1), (4,10,1).\\
        \*A_7:& (5,8,1), (6,9,1), (7,10,1).
    \end{array}
\end{equation}

For the objective $\*C$, we use the Kronecker product:
\begin{equation}
    \*{\bar B}_i^T \*R^\T \*{\bar y}_i = (\*{\bar y}_i \otimes \*{\bar B}_i)^\T \mathrm{vec}(\*R^\T).
\end{equation}

The $\*R^\T$ term can be permuted to $\*R$ by rearranging elements: $\mathrm{vec}(\*R^\T) = \*P\mathrm{vec}(\*R)$ for appropriate permutation matrix $\*P$. Expanding the objective and dropping constant terms, we have:
\begin{equation}
    \*C = \begin{bmatrix}
        \*F & \*g \\
        \*g^\T & 0
    \end{bmatrix}\!,
\end{equation}
where
\begin{equation}
    \*F \triangleq \*K^\T\*C_1\*{\hat B}^2\*C_1 \*K - 2\*K^\T\*C_1\*K + \lambda\*K^\T\*C_1\*C_1\*K,
\end{equation}
\begin{equation}
    \*g \triangleq \*K^\T\*C_1\*{\hat B}^2\*c_2 -\*K^\T\*c_2 + \lambda\*K^\T\*C_1\*c_2,
\end{equation}
\begin{equation}
    \*K \triangleq \sum_{i=1}^N (\*{\bar y}_i \otimes \*{\bar B}_i)^\T \*P.
\end{equation}

\section{Gauss-Newton and Levenberg-Marquardt Solvers}
\label{appendix:scf_gn}

In this section we derive the Gauss-Newton (\textsf{G-N}) and Levenberg-Marquardt (\textsf{L-M}) solvers for~\eqref{eq:paceqcqp_rotonly}. We follow the derivation of Gauss-Newton for nonlinear optimization outlined in~\cite{Barfoot17book}. This amounts to linearizing rotations and writing the Jacobians.

We use the axis-angle representation of rotations and linearize about a small perturbation. Let $\'\omega\in\RR^3$ be the rotation by angle $\|\'\omega\|_2$ about the axis given by the direction of $\'\omega$. Define the hat operator to take the skew-symmetric form of $\'\omega$. To convert this representation into the matrix representation, use a matrix exponential:
\begin{equation}
    \*R(\'\omega) = \exp({\'{\hat\omega}}).
\end{equation}

For a small axis-angle perturbation $\'\delta$, we use a first-order approximation of the rotation $\*R(\'\delta) \approx \eye_3 + \'{\hat\delta}$. Thus, for initial rotation $\*R_0$, the perturbed rotation is:
\begin{equation}
    \*R = \*R(\'\delta)\*R_0 \approx \*R_0 + \'{\hat\delta}\*R_0.
\end{equation}

We are now ready to linearize the optimization process and write the Gauss-Newton update step. Rewriting eq.~\eqref{eq:paceqcqp_rotonly} in terms of residuals, we have the following optimization problem:
\begin{equation}
    \min_{\*R\in\SO3} \sum_{i=1}^N \|\*r_i(\*R)\|^2 + \|\*r_c(\*R)\|^2.
\end{equation}

Gauss-Newton updates $\'\delta$ by the following quantity (see~\cite{Barfoot17book} for a derivation):
\begin{equation}
    \label{eq:GNupdate}
    -\*H^{-1} 
    \left( 
        \*J_c(\*R_0)^\T\*r_c(\*R_0) + 
        \sum_{i=1}^N \*J_i(\*R_0)^\T\*r_i(\*R_0)
    \right),
\end{equation}
where
\begin{equation}
    \*H\triangleq
    \*J_c(\*R_0)^\T\*J_c(\*R_0) + \sum_{i=1}^N \*J_i(\*R_0)^\T\*J_i(\*R_0) + \lambda\eye_3.
\end{equation}

In eq.~\eqref{eq:GNupdate} the $\*J$ variables correspond to the Jacobians to their corresponding residuals. Picking $\lambda=0$ corresponds to Gauss-Newton. Levenberg-Marquardt is identical for $\lambda > 0$, which primarily helps reguarlize the inverse. Update $\*R_0$ by $\*R(\'\delta)\*R_0$ at each iteration. It remains only to derive the Jacobians.

For each residual, the corresponding Jacobian satisfies $\*r(\*R_0 + \'{\hat\delta}\*R_0) = \*r(\*R_0) + \*J(\*R_0)\'{\hat\delta}$. All the residuals are linear in $\*R$, so no further approximations are needed. We use the identity $\*{\hat a} \*b = -\*{\hat b} \*a$ for any $\*a, \*b\in\RR^3$. We omit the algebra, which consists of manipulating the residuals into the form above using the identity. The expressions for the Jacobians are given below:

\begin{equation}
    \*J_i(\*R) = \*R^\T \*{\hat{\bar y}}_i - 2 \*{\bar B}_i\*C_1\sum_{j=1}^N \*{\bar B}_j^\T \*R^\T \*{\hat{\bar y}}_j,
\end{equation}
\begin{equation}
    \*J_c(\*R) = 2\sqrt{\lambda} \*C_1 \sum_{j=1}^N \*{\bar B}_j^\T \*R^\T \*{\hat{\bar y}}_j.
\end{equation}

\section{Extra Experimental Results with Larger Shape Library}
\label{appendix:scf_extraresults}
Performance with a larger shape library strongly depends on the choice of regularization constant $\lambda$. For $\lambda\rightarrow 0$ the problem becomes ill-conditioned and all solvers slow significantly. Large $\lambda$ greatly reduces shape ambiguity by imposing a strong prior. For these results, we test with $K=25$, $N=10$, and $\lambda=1.0$. This moderate choice of $\lambda$ gives similar results to~\prettyref{sec:scf}.

\begin{figure}[t]
    \centerline{\includegraphics[width=\linewidth]{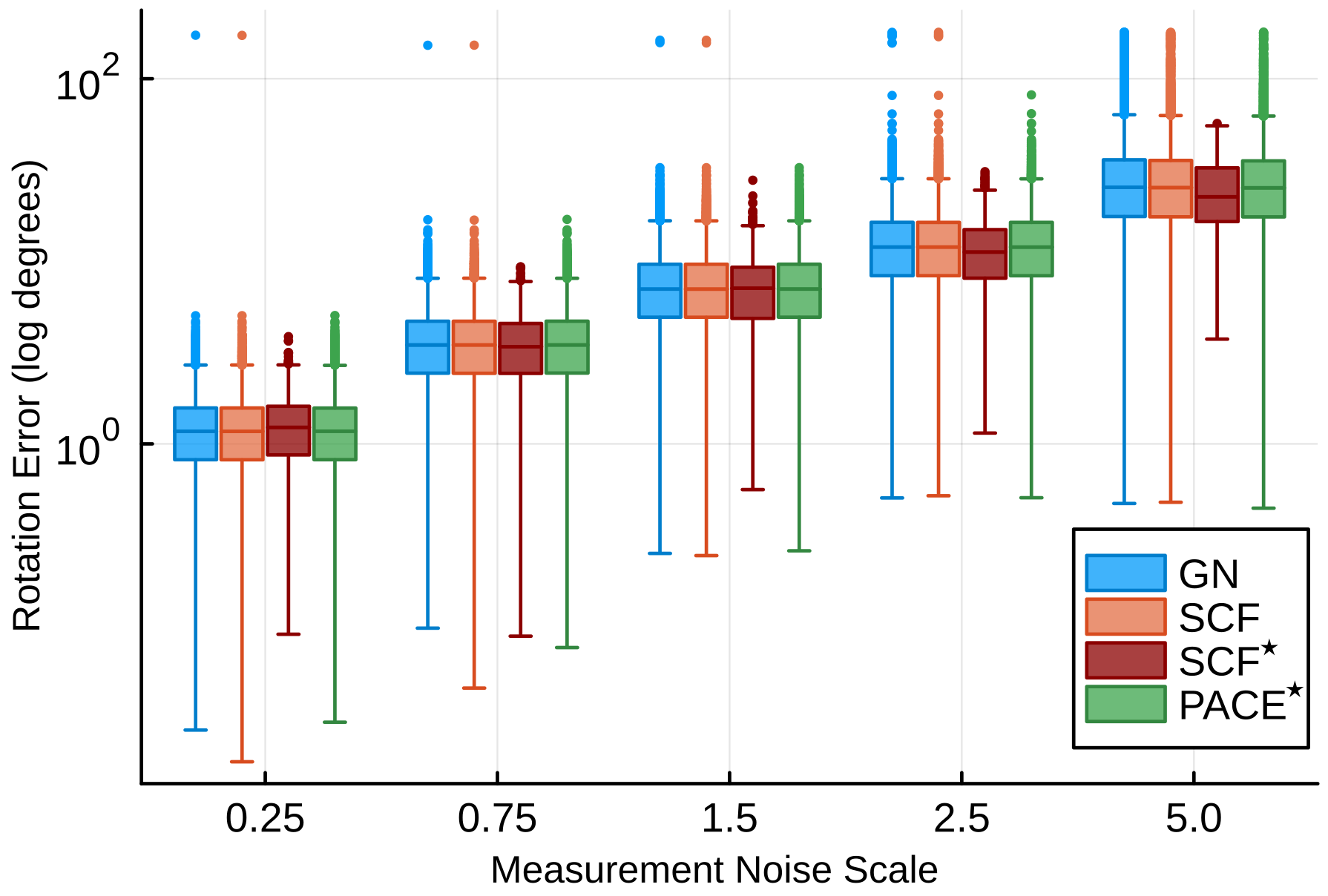}}
    \caption{\textbf{Distribution of Rotation Errors for Larger Shape Library.} For $K>N$ the performance depends heavily on choice of regularization $\lambda$. For $\lambda=1.0$, GN, SCF, and PACE have very similar rotation accuracy across noise scales. \scfstar~and \pace~show only the globally optimal estimates.
    \vspace{-10pt}
    }
    \label{fig:scf_mnoise_highK}
\end{figure}

\begin{table}[tb]
    \centering
    \caption{Runtimes on Synthetic Dataset (Large $K$)\vspace{-10pt}}
    \label{tab:scf_runtimes2}
    \adjustbox{width=\linewidth}
    {\begin{tabular}{rcccc}
        \toprule
                 & \multicolumn{2}{c}{$\sigma_m =0.25$} & \multicolumn{2}{c}{$\sigma_m = 2.5$} \\
                \cmidrule(lr){2-3}\cmidrule(lr){4-5}
                \textbf{Method}
                 & Mean (ms)    & p90 (ms)   & Mean (ms)  & p90 (ms)   \\ \midrule
        SCF &        \textbf{0.207} &      \textbf{0.207} &      \textbf{0.234} &     \textbf{0.248} \\ 
            GN &        0.487 &      0.545 &      0.643 &     0.776 \\ 
            LM &        0.494 &      0.557 &      0.642 &     0.777 \\ 
        Manopt &        1.118 &      1.281 &      1.129 &     1.323 \\ \midrule
        \scfstar &        0.244 &      0.241 &      0.267 &     0.277 \\
        \pace &        1.712 &      1.707 &      1.738 &     1.747 \\ 
        \bottomrule
    \end{tabular}}
\end{table}

We compare the estimation accuracy of GN, SCF, \scfstar, and \pace~in \prettyref{fig:scf_mnoise_highK}. As before, the methods achieve virtually the same performance across noise scales, while the global optimality certificate filters out outlier results for \scfstar. Perhaps unsurprisingly, the large $K$ case is a more difficult problem. While \pace~reports an optimality certificate for every estimate, \scfstar~only certifies optimality for $11.5$\%, $10.8$\%, $8.8$\%, $6.2$\%, and $1.4$\% of estimates at increasingly noise scales. The runtimes at high $K$, in~\prettyref{tab:scf_runtimes2}, are slightly longer but have the same trends as before. Interestingly, \pace~suffers the most modest increase in runtime. Although it is only $2-3$ times faster than G-N, SCF is conclusively the fastest method.

\end{document}